\documentclass[runningheads]{llncs}
\usepackage{amsmath,amssymb}

\usepackage{float}
\usepackage{amsmath,amssymb}
\usepackage{caption} 
\usepackage[utf8]{inputenc}
\usepackage[english]{babel}

\usepackage{graphicx}
\usepackage{sectsty}

\sectionfont{\fontsize{12}{15}\selectfont}
\subsectionfont{\fontsize{11}{13}\selectfont}

\usepackage{wrapfig}

\usepackage{graphicx}

\usepackage[utf8]{inputenc}
\usepackage[T1]{fontenc}

\begin{document}
\title{ Augmenting Physiological Time Series Data: A Case Study for 
Sleep Apnea Detection}
\titlerunning{   Augmenting Physiological Time Series Data}
%
\author{Konstantinos Nikolaidis\inst{1}
\and
Stein Kristiansen\inst{1} \and
Vera Goebel\inst{1} \and
Thomas Plagemann\inst{1} \and
Knut Liestøl\inst{1} \and
Mohan Kankanhalli\inst{2}}
\authorrunning{K. Nikolaidis et al.}
\institute{Department of Informatics, University of Oslo, Gaustadalleen 
23B, 0316 Oslo, Norway \and
Department of Computer Science, National University of Singapore,   COM1, 
13 Computing Drive, 117417, Singapore \\
}
\maketitle              
\begin{abstract}
Supervised machine learning applications in the health domain often face 
the problem of insufficient training datasets. The quantity of labelled 
data is small due to privacy concerns and the cost of data acquisition 
and labelling by a medical expert. Furthermore, it is quite common that 
collected data are unbalanced and getting enough data to personalize 
models for individuals is very expensive or even infeasible. This paper 
addresses these problems by (1) designing a recurrent Generative 
Adversarial Network to generate realistic synthetic data and to augment 
the original dataset, (2) enabling the generation of balanced datasets 
based on heavily unbalanced dataset, and (3) to control the data 
generation in such a way that the generated data resembles data from 
specific individuals. We apply these solutions for sleep apnea detection 
and study in the evaluation the performance of four well-known 
techniques, i.e., K-Nearest Neighbour, Random Forest, Multi-Layer 
Perceptron, and Support Vector Machine. All classifiers exhibit in the 
experiments a consistent increase in sensitivity and a kappa statistic
increase by between 0.72$\cdot 10^{-2}$ and 18.2$\cdot 10^{-2}$.
\end{abstract}
\section{Introduction}
\label{Intro}
The development of deep learning has led in recent years to a wide range 
of machine learning (ML) applications targeting different aspects of 
health \cite{ravi2017deep}. Together with the recent development of 
consumer electronics and physiological sensors this promises low cost 
solutions for health monitoring and disease detection for a very broad 
part of the population at any location and any time. The benefits of 
automatic disease detection and especially early prognosis and life 
style support to keep healthy are obvious and result in a healthier 
society and substantial reduction of health expenses. However, there are 
high demands on the reliability of any kind of health applications and 
the applied ML methods must be able to learn reliably and operate with 
high performance. To achieve this with supervised learning, appropriate 
(labelled) datasets gathered with the physiological sensors that shall 
be used in a health application are needed for training such that 
classifiers can learn to sufficiently generalize to new data. However, 
there are several challenges related to training datasets for health 
applications including data quantity, class imbalance, and 
personalization.
  \par In many domains, the quantity of labelled data has increased 
substantially, like computer vision and natural language processing, but 
it remains an inherent problem in the health domain  
\cite{ravi2017deep}. This is due to privacy concerns as well as the 
costs of data acquisition and data labelling. Medical experts are needed 
to label data and crowdsourcing is not an option. To enable medical 
experts to label data, data are typically acquired with two sensor sets. 
One set with the sensors that should be used in a health application and 
one sensor set that represents the gold standard for the given task. 
This problem is magnified by the fact that any new physiological sensor 
requires new data acquisition and labelling. Furthermore, there is a 
high probability that the data acquisition results in an unbalanced 
dataset.  Since many health applications aim to detect events that 
indicate a health issue there should “ideally” be equally many time 
periods with and without these events. In general, this is unrealistic 
for a recording from an individual as well as across a larger population 
that is not selected with prior knowledge of their health issues. For 
example, in the recent A3 study \cite{traaen2019treatment}
 at the Oslo 
University Hospital individuals with atrial fibrillation were screened 
for sleep apnea. In a snapshot from this study with  328 individuals, 62 
are classified as normal, 128 with mild apnea, 100 with moderate apnea, 
and 38 with severe apnea. The severeness of sleep apnea is captured by the Apnea Hypopnea Index (AHI) which measures the average number of apnea events per hour and is classified as follows: AHI$<$15, (normal), 15$\leq$ AHI$<$30, (moderate), AHI$\geq$30, (severe)\footnote{From a ML viewpoint only individuals with severe sleep apnea would produce balanced recordings}.   It is unrealistic to expect that a 
sufficiently large dataset for training can be collected from each 
individual, because it is inconvenient, requires medical experts to 
label the data, and might be infeasible due to practical reasons for 
those that develop the application and classifier.

   \par The objectives of this work are to address these problems with 
insufficient datasets in the health domain: (1) generate synthetic 
data from a distribution that approximates the true data distribution to 
enhance the original dataset; (2) use this approximate distribution to 
generate data in order to rebalance the original dataset;  (3) 
examine the possibility to generate personalized data that correspond to 
specific individuals; and (4) investigate how these methods can lead to  performance improvements for the classification task. 

  \par The mentioned problems are relevant for many applications in the 
health domain. As a proof-of-concept, we focus in our experimental work 
on the detection of obstructive sleep apnea (OSA). OSA is a condition 
that is characterized by frequent episodes of upper airway collapse 
during sleep, and is being recognized as a risk factor for several 
clinical consequences, including hypertension and cardiovascular 
disease. The detection and diagnosis is performed via polysomnography 
(PSG). PSG is a cumbersome, intrusive and expensive procedure with very 
long waiting times. Traditionally, PSG is performed in a sleep 
laboratory. It requires the patient to stay overnight and record various 
physiological signals during sleep, such as the electrocardiogram, 
electroencephalogram, oxygen saturation, heart rate,  and respiration 
from the abdomen, chest and nose. These signals are manually evaluated 
by a sleep technician to give a diagnosis.  In our earlier work 
\cite{kristiansen2018data}, we could show that machine learning can be 
used to classify PSG data with good performance, even if only a subset 
of the signals is used, and that the quality of collected data with 
commercial-of-the-shelf respiratory sensors (like the Flow sensor from 
Sweetzpot costing approximately 200 Euro) approaches the quality of 
equipment used for clinical diagnosis \cite{loberg2018quantifying}.
  \par  In this work, we use different  conditional recurrent GAN designs,  and  four well known classification techniques, i.e., K-Nearest Neighbor (KNN), Random Forest (RF), Multi-Layer Perceptron (MLP), and Support Vector Machine (SVM)  to achieve the aforementioned objectives.  Since we want to use datasets that are publicly available and open access,  we use the  Apnea-ECG and  MIT-BIH databases from Physionet \cite{ApneaEcg,MITBIH} for our experiments. The reminder of this paper is organized as follows: In Section 2 we  examine related works. We present our methods in Section 3. In Section 4 
we evaluate these methods by performing three experiments. Section 5  concludes this paper.

\section{Related Work}
\label{Related_Work}
Although the GAN framework \cite{goodfellow2014generative} has recently 
acquired significant attention for its capability to generate realistic 
looking images 
\cite{radford2015unsupervised,isola2017image}, we are 
interested in time series generation. The GAN is not as widely used for 
time series generation as for images or videos, however, several works 
which investigate this approach exist  \cite{mogren2016c}. There are, also 
relevant applications for sequential discrete data 
\cite{yu2017seqgan}.
\par In relation to our objectives most works are related to Objective 1 
  \cite{esteban2017real,choi2017generating}.  Hyland et al.  \cite{esteban2017real} use a conditional recurrent GAN 
to generate realistic looking intensive care unit data, which have 
continuous time series form. They use a conditional recurrent GAN (based 
on \cite{mirza2014conditional}), to generate data preconditioned on 
class labels. Among other experiments, they train a classifier to 
identify a held out set of real data and show the possibility of 
training exclusively in synthetic data for this task. They also 
introduce the opposite procedure (train with the real data and test on 
the synthetic) for distribution evaluation. We use similar methods to 
synthesize data in the context of OSA, but we expand these techniques 
by introducing a metric for evaluating the synthetic data quality which 
is based on their combination. We also investigate methods to give different 
importance to different recordings.
  Other works related to 
medical applications of GANs include \cite{hwang2017disease} and 
\cite{che2017boosting}. Our work is  associated with the use of multiple 
GANs in combination and uses different design and metrics from the above 
works (both works use designs based on combinations of an auto-encoder 
and a GAN). Many approaches that include multiple GANs exist such 
as \cite{durugkar2016generative,hoang2017multi}.

\par We note that most of the related work with the exception of 
\cite{che2017boosting} focuses individually on the synthetic data 
generation and evaluation, and not how to use these data to augment the 
original dataset to potentially improve the generalization capability of 
other classifiers. To the best of our knowledge only few works 
\cite{douzas2018effective,rezaei2019recurrent,mariani2018bagan} exist 
that examine the potential application of GANs to produce realistic 
synthetic data for class rebalancing of a training dataset. Only one of 
them uses specifically a recurrent GAN architecture. Finally, we did not  find any relevant work that depicts the data distribution  as a mixture of different  recording distributions, with the end-goal of producing more personalized synthetic data.

\section{ Method}
\label{method}

The goal of data augmentation in this work is to train classifiers to 
successfully detect in physiological time series data health events of 
interest. In our use case this means to classify every 30 or 60 second 
window of a sleep recording as apneic (i.e., an apnea event happened) or 
non-apneic.
\begin{figure*}
\vskip -0.5cm
\raggedright
\includegraphics[width=\textwidth,height=6cm]{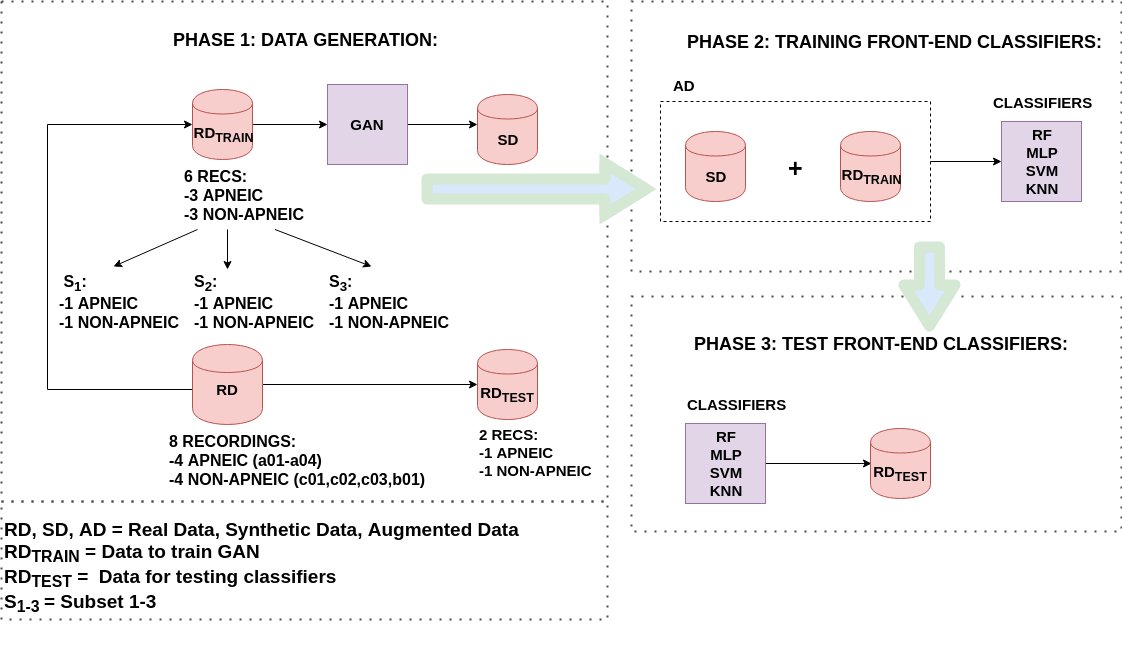}
\vskip -0.6cm
\caption{GAN Augmentation} \label{fig1}
\vskip -0.6cm
\end{figure*}

\par Our approach is based on a conditional recurrent GAN to generate a 
synthetic dataset (SD, see Figure \ref{fig1}) to augment the original 
training dataset (RD$_{TRAIN}$)  (Objective 1) and to rebalance an 
unbalanced RD$_{TRAIN}$ (Objective 2). Furthermore, we extend the single 
GAN architecture to a multiple GAN architecture to generate more 
synthetic data that is potentially closer to the test data to enable 
personalized training (Objective 3). In this section, we introduce the datasets we use, the two GAN architectures, and the metrics used to evaluate the quality of the generated data. 

\subsection{Data}
\label{eval_Data}

In this work  we focus on the nasal airflow signal (NAF), because it can
adequately be used to  train a classifier to recognize apneas and yields the best single signal performance
as shown in our previous work \cite{kristiansen2018data}.  Furthermore, NAF is contained in most   recordings (in 12 recordings\footnote{slp01, slp02a , slp02b, slp03 , slp04, slp14, slp16, slp32, slp37, slp48, slp59, slp66, slp67x}) in the MIT-BIH database. From the  Apnea-ECG database we use the eight sleep recordings (i.e., a01, a02, a03, a04, c01, c02, c03,
b01) that contain the NAF signal with durations  7- 10 hours. From MIT-BIH we use the 12 recordings that include NAF signal. Note that  MIT-BIH has low  data quality  (noisy wave-forms, values out of bounds, etc), especially when compared to Apnea-ECG. 
\par The sampling frequency is 100Hz for Apnea-ECG and 250Hz for MIT-BIH and all
recordings contain labels for every minute window of breathing for Apnea-ECG and for every 30 seconds window for MIT-BIH. These labels classify a
window as apneic  and non-apneic.  For  Apnea-ECG,  half of the 8 recordings  are classified as severe OSA (a01-a04,  called "apneic" recordings) and half are classified as normal OSA (c01-c03,b01,  called  "non-apneic").  AHIs vary from 0 to 77.4.   For MIT-BIH,  AHIs vary from 0.7 to 100.8. The only preprocessing we perform is rescaling and downsampling the data to 1Hz.

\subsection{Single GAN Architecture}

In order to solve  the problems of too small and unbalanced  dataset we generate synthetic data and augment the original dataset. Due to its recent successes in generating realistic looking 
synthetic data e.g. images and music, we use the GAN framework to 
produce realistic looking synthetic time series data. In particular, we 
use a conditional recurrent GAN. The conditional aspect allows us to 
control the class of the generated data (apneic, non-apneic). Thus, data 
from both classes can be generated and the front-end classifiers are 
able to learn both apneic and non-apneic event types. The generative 
network G() takes as input random sequence from a distribution $p_z(z)$  and returns a 
sequence that after training should resemble our real data. The 
discriminator D() takes as input the real data with distribution 
$p_{Data}(x)$ and the synthetic data from G, and outputs the probability 
of the input being real data. Using cross-entropy error, we  obtain 
the value function \cite{goodfellow2014generative}:
\begin{equation}
\min_G \max_D V(D,G) = \mathbb{E}_{x\sim p_{Data}(x)}[\log 
D(x)]+\mathbb{E}_{z\sim
p_{Z}(z)}[1- \log D(G(z))]
\label{eq1}
\end{equation}
\par G has the objective to minimize the probability that D correctly 
identifies the generated data as synthetic (see the 
second term of Eq. \ref{eq1}). D has the objective to maximize the 
probability to correctly classify data as either real or synthetic.
\par The objective of the generator is to fool the discriminator such 
that it classifies generated data as real. Through the training the 
generator learns to produce realistic looking synthetic data. 
Consequently, the generated data distribution converges to the real data 
distribution  \cite{goodfellow2014generative}. Inspired by 
\cite{esteban2017real}, we use a conditional LSTM 
\cite{hochreiter1997long} as G and D, because we are interested in time 
series generation of sequentially correlated data. LSTMs are able to 
store information over extended time intervals and avoid the vanishing 
and exploding gradient issues  \cite{goodfellow2016deep}. G produces a 
synthetic sequence of values for the nasal airflow and D classifies each 
individual sample as real or fake based on the history of the sequence.

\subsection{ Multiple GAN Architecture}
\label{meth_exp3}

\par The aim for this approach is to ensure that the SD represents in a 
realistic manner all recordings in RD$_{TRAIN}$.
 Each person, depending 
on various environmental and personal factors has different breathing 
patterns.

\begin{wrapfigure}{r}{0.5\textwidth}
  \vskip -0.5cm
  \begin{center}
    \includegraphics[width=0.48\textwidth,height=5cm]{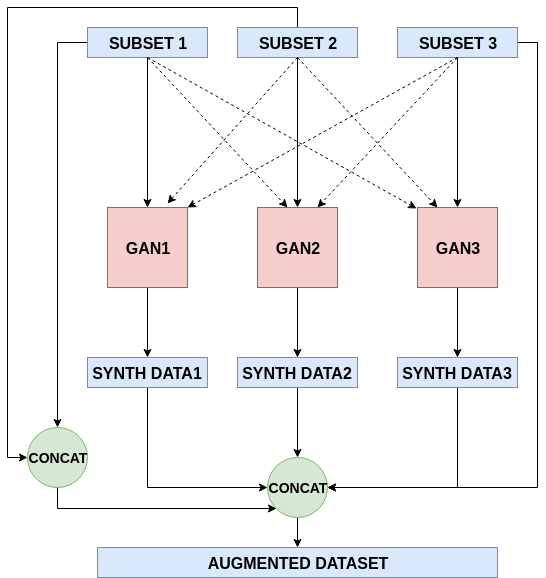}
  \end{center}
  \vskip -0.5cm
  \caption{Three GANs trained separately with a chance to 
interchane subsets.}
\vskip -0.5cm
\label{fig:Exp3_1}
\end{wrapfigure}
 Common general patterns exist among different people depending 
on different factors of different recordings, but individual 
characterization is possible.
Even for the same person, the recordings  of different sessions can be different. These changes are often 
described as bias towards a particular patient  
\cite{goodfellow2016deep}. We follow a different approach and make the 
hypothesis that different recording sessions have different data 
distributions, which together constitute the total apnea/non-apnea 
distribution of the dataset. In our case different recordings correspond 
to different individuals.
A distinction is made between the recordings  and the modes in their distribution since a recording can have more than 
one mode in its distribution, and different modes in the feature space 
can be common for different recordings. Since we have insufficient data 
per recording to successfully perform the experiments of this section, 
we define disjoint subsets of recordings (hereby called 
\textit{subsets}), the union of which constitutes the original recording 
set.
Under this hypothesis, the data distribution can be depicted as a 
mixture of the different recording distributions:

\begin{equation}
p_{Data}(x)= \sum^{k_{rec}}_{i=0} w_{r_i}p_{rec_i}(x)= 
\sum^{k_{sub}}_{j=0} w_{s_j}p_{sub_j}(x)
\end{equation}
\vskip -0.3cm

with:
\vskip -0.3cm

\begin{equation}
p_{sub_j}(x)= \sum_{l\in sub_j} w_{sb_lj}p_{rec_l}(x)
\end{equation}

where $k_{rec}$ is the total number of recordings, $k_{sub}$ is the 
total number of subsets,  $p_{rec_i}$ is the data distribution of 
recording i, and $w_{r_i}=1/k_{rec}$ assuming equal contribution per 
recording, $p_{sub_j}$ and  $w_{s_j}$ is the distribution and weights of 
subset j, and $w_{{sb}_lj}$ the weights of the recording within each 
subset.

We restate Eq. \ref{eq1} to explicitly include the distributions of the 
subsets by dedicating a pair of G and D to each subset.  This allows 
each GAN to prioritize the data from its respective subset, thus making 
it less probable to exhibit mode collapse for modes contained in the 
examined recordings. Each subset contains one apneic and one non-apneic 
recording (see Section 3.1, 4.4).
\par The goal of this method is to properly represent all recordings in 
the SD. The potential decrease of collapsing modes due the use of 
multiple GANs for different data is an added benefit. There are relevant 
publications that use similar ensemble techniques to specifically 
address this issue backed by theoretical or methodological guarantees 
\cite{tolstikhin2017adagan,hoang2017multi}.

\par Since the amount of data per recording is too low to train GAN with 
only two recordings, we allow each GAN to train with data from the 
training subset of another GAN with a controllable probability  (see Figure \ref{fig:Exp3_1}).  Per iteration, for 
GANj we perform a weighted dice toss such that 
$J=(1,2...,j,...,k_{sub})$, and $ 
\mathbf{p}=(p_1,p_2,...p_j,...p_{k_{sub}})$ where J is a random variable 
following the multinomial distribution and $\mathbf{p}$  the parameter 
probability vector of the outcomes. For GANj $p_j=p$, and 
$p_1=p_2=...=p_i..=p_{k_{sub}}=\frac{1-p}{k_{sub}-1}\forall i\neq j$ for 
a chosen value $p$ . Note that the larger the chosen $p$, the more 
pronounced the modes of the recording combination that corresponds to 
GANi will be. It is relatively straightforward to show that:

\begin{proposition}
A GAN satisfying the conditions of Proposition 2 of 
\cite{goodfellow2014generative} and  trained with a dataset produced 
from the above method will converge to the mixture  distribution: 
$p_s(\mathbf{x})=\sum_i^{k_{sub}} w_ip_{sub_i}(\mathbf{x})$ where $w_i= 
P(J=j)$.
\end{proposition}

Based on this proposition, this method creates a variation of the 
original dataset, that gives different predefined importance to the 
different subsets (see Appendix for details). The same proposition holds 
for individual recordings. The value function now for a GAN takes the 
following form:
  \begin{equation}
\min_G \max_D V(D,G) = \mathbb{E}_{x\sim p_s(x)}[\log 
D(x)]+\mathbb{E}_{z\sim
p_{Z}(z)}[1- \log D(G(z))]
\label{gan_value_mixture}
\end{equation}

\subsection{Metrics}

\par Measuring the quality of data produced by a GAN is a difficult 
task, since the definition of “realistic” data is inherently vague. 
However, it is necessary, because the performance of the front-end 
classifiers is not necessarily a direct measurement of how realistic the 
synthetic data are. In this subsection we introduce the metrics we 
use to measure the quality of the synthetic data.
\subsubsection{T metric:}
\par    Hyland et al. \cite{esteban2017real}  introduce two empirical evaluation metrics for data quality: TSTR (Train on Synthetic Test on Real) and TRTS (Train on Real Test on Synthetic). Empirical evaluation indicates that these metrics are useful in our case, however each one has  disadvantages. To solve some of these issues we combine them  via taking their harmonic mean (in the Appendix we explain problems with these metrics and reasons to use the harmonic mean):
\begin{equation}
T=\frac{2*TSTR*TRTS}{TSTR+TRTS}
\end{equation}

\subsubsection{ MMD:}
\par We chose the Maximum Mean Discrepancy (MMD) 
\cite{gretton2007kernel} measurement since other well-established 
measurements (e.g., log likelihood) are either not well suited for GAN 
assessment, because plausible samples do not necessarily imply high log 
likelihood and vice versa \cite{theis2015note}, or they are focused on 
images, like the inception score \cite{salimans2016improved} and the 
Frechet Inception distance. There is also a wide variety of alternative 
approaches  \cite{borji2019pros}, however we use the MMD since it 
is simple to calculate, and is generally in line with our visual 
assessment of the quality of the generated data.
\par We follow the method from \cite{sutherland2016generative} to 
optimize the applied MMD via maximizing the ratio between the MMD 
estimator and the square root of the estimator of the asymptotic 
variance of the MMD estimator (the t-statistic). Inspired by 
\cite{esteban2017real}, we further separate parts of the real and 
synthetic datasets to MMD training and MMD test sets (each contains half 
real and half synthetic data points). To maximize the estimator of the 
t-statistic for the training data we run gradient descent to the 
parameters of our kernel (i.e., Radial Basis Function (RBF) with 
variance $\sigma$ as parameter). Then we test the MMD measurement on the 
MMD test set with the parameters that have been optimized with the 
training set. In  the next section we  evaluate  the 
data based on these metrics.

\section{Evaluation}
In this section, we present the implementation and evaluation of our 
experiments. To analyze how well we can achieve our objectives with the 
two GAN architectures, we design three experiments. Before we describe 
these experiments and their results, we 
analyze in 
Section \ref{Quality_eval}  the quality of the synthetic data with the T-metric, the MMD, and 
visual inspection. In Sections 4.2-4.4 we present 
and analyze  the  experiments we conduct. 
Together with accuracy, specificity, and sensitivity we use the kappa 
coefficient \cite{cohen1960coefficient} as performance metric since it  better 
captures the performance of two-class classification in a single metric 
than accuracy. For all experiments, the pre-processing of the data is minimal (Section \ref{eval_Data}) and we use a wide variety of relatively basic methods as front-end classifiers. This is because we want to focus on investigating the viability of GAN augmentation as a means of performance improvement for a general baseline case. However,  the GAN augmentation is applicable to any type of data (e.g., pre-processed apnea data) and is independent of the front-end classifiers.
For details about the GAN and the front-end classifiers parameters 
and design please refer to Appendix.

\subsection{Data Quality Evaluation}
\label{Quality_eval}
\par
\begin{figure}
\vskip -0.5cm
\centering
  \includegraphics[scale=0.24]{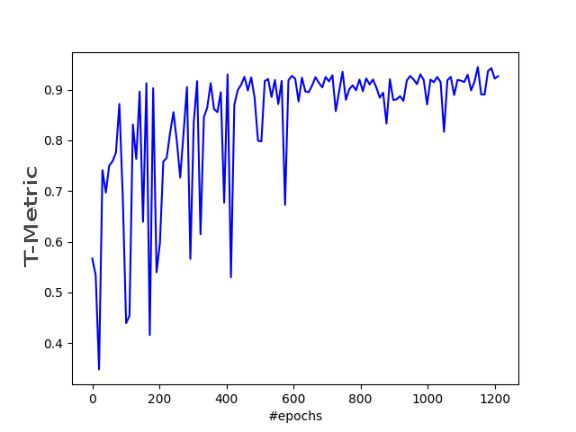}
  \label{fig:Q_sub1}
  \includegraphics[scale=0.3]{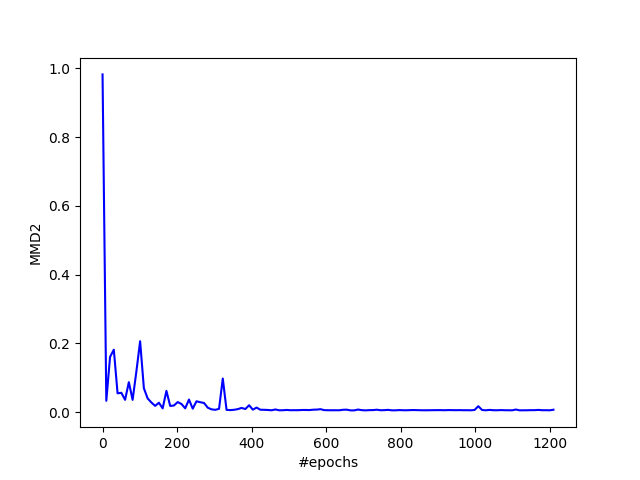}
   \label{fig:Q_sub2}
\caption{Mean of T-metric (left) and MMD (right) scores throughout the 
GAN training }
\label{fig:Quality_Mes}
\vskip -0.25cm
\end{figure}

 To measure the similarity between the synthetic and the 
real distribution we use the MMD and T metrics (see example in Figure 
\ref{fig:Quality_Mes}). We execute the tests every 10 epochs during 
training. Both scores improve as the training procedure progresses, 
until they stabilize (with minor variation). The T metric is more 
unstable with epochs with high score in the initial training phase. 
However,

\begin{figure}
\centering
  \includegraphics[scale=0.225]{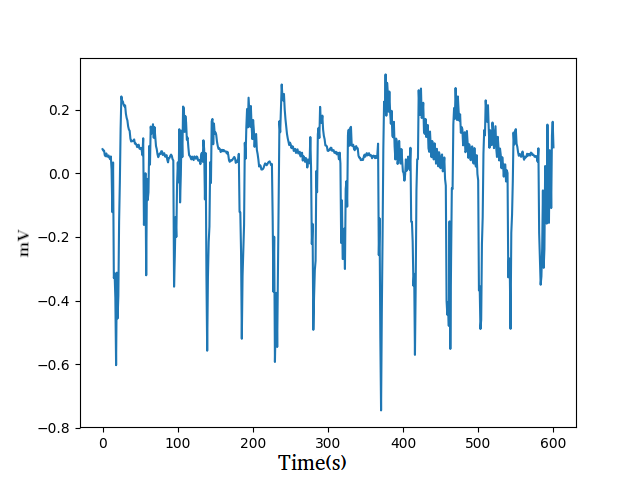}
  \label{fig:Real1}
  \includegraphics[scale=0.225]{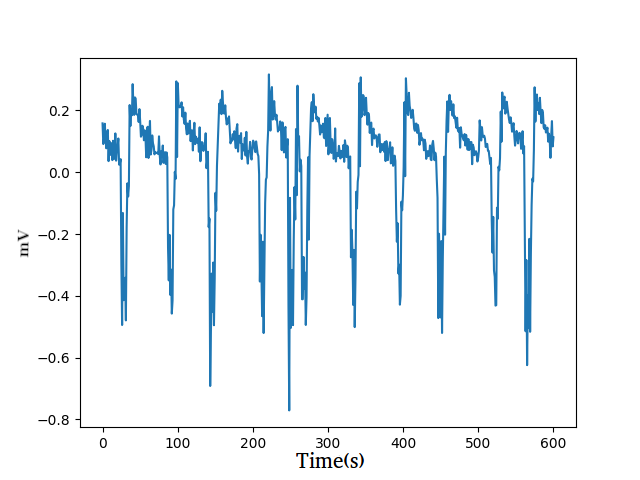}
  \label{fig:Synth1}
\caption{Real apneic data (left) and good synthetic
apneic data (right) for 600sec}
\label{fig:Apnea_im}
\vskip -0.5cm
\end{figure}
after epoch 600, the 
performance of the metric stabilizes around 0.9. Similarly, the majority 
of MMD variations stop (with few exceptions) around epoch 400.
\par
\begin{figure}
\vskip -0.5cm
\centering

  \includegraphics[scale=0.225]{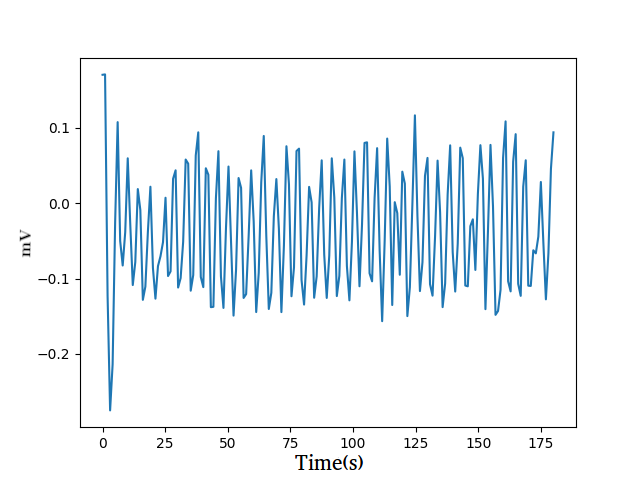}
  \label{fig:Non_apnea_Real1}
  \includegraphics[scale=0.225]{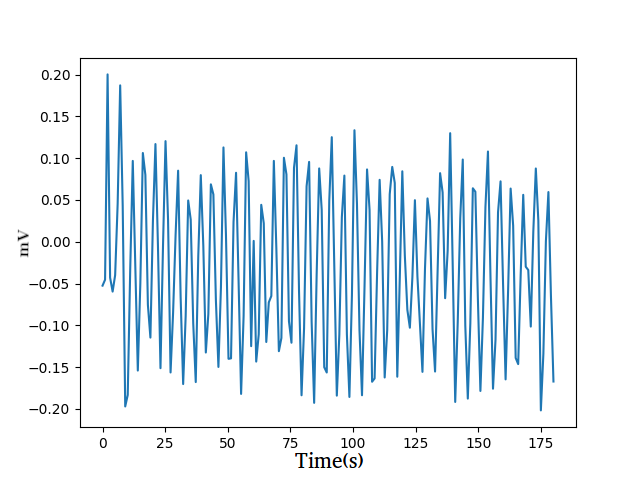}
  \label{fig:Non_apnea_synth1}

\caption{Real  (left) and good synthetic (right)
non-apneic data ,  175 sec}
\label{fig:Non_apnea_im}
\vskip -0.5cm
\end{figure}

 Another important criterion for recognizing whether the generated 
data are realistic is the visual inspection of the data. Although not as 
straightforward as for images, apnea and non-apnea data can be visually 
distinguished. In Figures \ref{fig:Apnea_im} and \ref{fig:Non_apnea_im} we 
show  examples of real and realistic-looking synthetic data.  The generated data are  realistic-looking and difficult to distinguish from the real.

\subsection{Experiment1: Data Augmentation}
\label{Exp1_eval}

In this experiment we investigate whether augmenting RD$_{TRAIN}$ with 
realistic SD generated from a GAN trained with the same RD$_{TRAIN}$ can 
have a positive impact on the front-end classifier performance.
\par \textbf{Experiment Description:} We iterate the following 
experiment 15 times for Apnea-ECG and 10 times for MIT-BIH: We partition 
RD into  RD$_{TRAIN}$ (with  50\% of RD data points),  RD$_{TEST}$ 
(25\%) and a validation set  (25\%) via random subsampling. With 
RD$_{TRAIN}$ we train GAN. The GAN training is very unstable for the 
data of the two datasets (especially for MIT-BIH), and a good quality 
based on our metrics and visual inspection does not necessarily 
correspond to high performance of the front-end classifiers. For this 
reason, we use the validation dataset to evaluate the front-end 
classifier performance. We save the trained GAN model periodically 
throughout training, generate SD, augment RD$_{TRAIN}$, and measure the 
front-end classifier performance on the validation set. The GAN with the 
maximum validation set performance, and empirically acceptable MMD and 
T-metric values is chosen to generate SD.

\par \textbf{Results:} Due 
to limited space we present in the main text only the kappa statistic 
for all front-end classifiers (Table \ref{table1}) , in addition to the accuracy sensitivity 
and specificity for the MLP classifier (Table \ref{table2}) to indicate the general behaviour 
we observe for all the classifiers. For accuracy, specificity, 
sensitivity for KNN, RF and MLP please refer to Appendix A. We use this 
presentation convention for all experiments.

  \begin{table}
  \vskip -0.3cm
  \caption{Kappa statistic and standard error for all the front-end classifiers for 
Apnea-ECG and MIT-BIH.All kappa values are multiplied by 100 for legibility}
\centering
\resizebox{10cm}{!}{%
\begin{tabular}{ | p{2.2cm} |p{1.9cm}|p{1.9cm}|p{1.9cm}|p{1.9cm}|}
\hline
\multicolumn{5}{|c|}{Kappa statistic (X$\cdot 10^{-2}$)   for Apnea-ECG (A), and  MIT-BIH (M)} \\
\hline
&MLP & RF &KNN & SVM \\
\hline
A: Baseline 
&85.89$ \pm$0.36&90.08$\pm$0.26&88.12$\pm$0.40&74.75$\pm$0.40\\
A: Exp1:Synth&78.29$\pm$0.97&83.88$\pm$0.56&85.76$\pm$0.49& 
75.04$\pm$0.55\\
A: Exp1:Augm &86.93$\pm$0.45&90.88$\pm$0.28&90.12$\pm$0.37& 
76.90$\pm$0.57\\
\hline
M: Baseline&25.04$\pm$0.88&30.95$\pm$1.10&27.15$\pm$1.01&0.0$\pm$0.0\\
M: 
Exp1:Synth&18.35$\pm$0.86&21.80$\pm$0.95&16.84$\pm$1.26&11.02$\pm$0.96 
\\
M: Exp1:Augm &27.01$\pm$0.61&33.01$\pm$0.87&29.22$\pm$1.01& 
14.93$\pm$1.22\\
\hline
\end{tabular}
}
\label{table1}
\vskip -0.3cm
\end{table}

\textit{Baseline} shows the performance of the front-end classifiers 
trained only with RD$_{TRAIN}$. For the synthetic case 
(\textit{Exp1:Synth}) they are trained only with SD, and for the 
augmented case (\textit{Exp1:Augm})  with RD$_{TRAIN}$ and SD.

For  Apnea-ECG,  Exp1:Augm exhibits  for all front-end classifiers 
 a statistically significant improvement of the mean of the 
kappa statistic at $p=0.05$. The p-value for the one-tailed two sample 
t-test  relative to the Baseline is: (MLP): p= 0.042, (RF): p=0.035, 
(KNN): p=0.005, (SVM): p=0.002. Notice that SD yields a good performance 
on its own, and even surpasses the performance of the Baseline for the 
SVM. We assume that this is due to the better balancing of the synthetic 
data in relation to the real. In SD, 50\% of the generated minutes are 
apneic and 50\% non-apneic, whereas in   RD$_{TRAIN}$   approximately 
62.2\% are non-apneic and 37.8\% are apneic depending on the random 
subsampling.

For MIT-BIH,  Exp1:Augm  shows a significant or nearly significant improvement of 
the kappa statistic values relative to the Baseline for all front-end classifiers when we perform the 
2-sample one tailed t-test, i.e., (MLP): p=0.012, (RF): p=0.062, (KNN): 
p=0.029, and (SVM): p$\simeq$0. The overall performance is very low, due 
to the very low data quality for this dataset. Since our pre-processing 
is minimal this is to be expected.  Notice that the SVM actually does 
not learn at all for the Baseline case. In all the iterations we 
performed it classifies all  minutes as non-apneic. Interestingly, 
both for Exp1:Synth and Exp1:Augm, there is a big improvement for the 
SVM, since  the algorithm successfully learns to a certain 
extent in these cases. We assume that this is due to the better class balance (more 
apneas present in the datasets of Exp1:Synth and Exp1:Augm). Generally, 
for MIT-BIH the augmentation seems to have a beneficial effect in 
performance.

  \begin{table}
  \vskip -0.3cm
  \caption{Accuracy specificity and  sensitivity for the MLP classifier }
\centering
\resizebox{8cm}{!}{%
\begin{tabular}{ | p{2.2cm} |p{1.9cm}|p{1.9cm}|p{1.9cm}|}
\hline
\multicolumn{4}{|c|}{MLP Classifier Apnea-ECG (A), and MIT-BIH (M)} \\
\hline
&Acc & Spec &Sens  \\
\hline
A: Baseline&93.19$\pm$0.17&94.78$\pm$0.19&90.83$\pm$0.39\\
A: Exp1:Synth&89.26$\pm$0.49&85.48$\pm$1.14& 95.02$\pm$0.94\\
A: Exp1:Augm &93.66$\pm$0.20&94.62$\pm$0.24& 92.28$\pm$0.46\\
\hline

M: Baseline&64.6$\pm$0.37&75.95$\pm$1.16&48.41$\pm$1.26\\
M: Exp1:Synth&59.76$\pm$0.5&61.6$\pm$2.58&57.17$\pm$3.16 \\
M: Exp1:Augm &64.7$\pm$0.25&69.92$\pm$0.78& 57.08$\pm$1.22\\
\hline
\end{tabular}
}
  \label{table2}
\vskip -0.3cm
\end{table}
 From Table \ref{table2} we notice that for Exp1:Augm,  the MLP  (both 
for MIT-BIH and Apnea-ECG) exhibits a clear improvement in sensitivity 
and  a small drop in specificity. This pattern is present for all 
front-end classifiers. For Exp1:Augm there is always a clear improvement 
in sensitivity, and either a small increase or decrease in specificity. 
This is an important advantage in a healthcare context since sensitivity 
reflects the ability of a classifier to recognize pathological events. 
This observation serves as a motivation for Experiment 2.

\par \textbf{Implications for OSA Detection:} The goal of this experiment is to reflect  a real  application scenario in which we have  relatively equal amount of  data from different patients to train with, and we perform classification for these  patients. An example could be mobile OSA detection for patients after monitoring. It  serves as an indication that augmentation with synthetic data can yield  performance improvements for classifiers that are trained with the goal of OSA detection.

\subsection{ Experiment2: Rebalancing Skewed Datasets}

To analyze how well the single GAN architecture can be used to 
rebalance a skewed dataset, Apnea-ECG needs to be modified, because it 
contains an equal number of apneic and non-apneic recordings (Section 
\ref{eval_Data}), and the apneic recordings contain many apneic minutes. 
Thus, the data are  lightly skewed towards non-apneic events in 
Apnea-ECG, with a ratio of 62.2\% non-apneic and 37.8\% apneic.
\par \textbf{Experiment Description:} We separate RD into RD$_{TRAIN}$ 
and RD$_{TEST}$  on a per-recording basis instead of a per event-basis 
as in the previous experiment. We randomly choose one apneic and one 
non-apneic recording as   RD$_{TEST}$ (i.e., a01 and b01 respectively), 
and as  RD$_{TRAIN}$ we use the remaining six recordings. We choose to 
evaluate this scenario using Apnea-ECG since it is the dataset for which 
our front-end classifiers exhibit the better 
performance.

\begin{figure}
  \vskip -0.5cm
\centering

  \includegraphics[width=6cm,height=1cm]{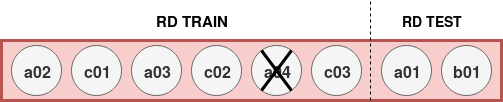}
\caption{Training and Test  sets for Experiment 2}
  \label{fig:Reb1}
  \vskip -0.5cm
\end{figure}
\par To create an unbalanced dataset, one apneic recording (i.e., a04 
chosen randomly) is removed from the training dataset RD$_{TRAIN}$ 
(Figure \ref{fig:Reb1}). Thus, the ratio is reduced to 72.2\% non-apneic 
27.8\% apneic when removing a04. The augmentation in this experiment 
rebalances the classes to 50\% apneic and 50\% non-apneic. This means 
that we only generate apneic data with the GAN (i.e., SD contains only 
apneic minutes) and combine them with the original dataset to form AD.

\begin{table}
\vskip -0.4cm
\caption{Kappa statistic and standard error for all front-end classifiers. }
\centering
\resizebox{10cm}{!}{%
\begin{tabular}{| p{2.1cm} |p{2cm}|p{2cm}|p{2cm}|p{2cm}|}
\hline
\multicolumn{5}{|c|}{Exp2: Kappa statistic  (X$\cdot 10^{-2}$)   a01b01-unbalanced} \\
\hline
&MLP & RF &KNN & SVM \\
\hline
Baseline&88.44$\pm$0.54  &91.92$\pm$0.26 &93.16$\pm$0.16 &74.6$\pm$0.2\\
Exp2:Augm&93.40$\pm$ 0.63 &94.56$\pm$0.16 &94.76$\pm$0.45
&92.88$\pm$0.64 \\
\hline
\end{tabular}
}
\label{table:ResUnbalanced}
\vskip -0.4cm
\end{table}

\par Note that a04 is removed from the training set both for the 
baseline/augmented training of the front-end classifiers and also for 
the training of the GAN, i.e., the apneic minute generation relies only 
on the other two apneic recordings. A validation set is extracted from 
a01b01. Throughout the training of the GAN the validation set is 
periodically evaluated by the front-end classifiers which are trained 
each time with AD. We choose the model that generates the SD with which 
the front-end classifiers perform the best on the validation set. For 
this experiment we perform 5 iterations.

\begin{table}
\vskip -0.4cm
\caption{ Accuracy, specificity and 
sensitivity for MLP }
\centering
\resizebox{8cm}{!}{%
\begin{tabular}{| p{2cm} |p{1.9cm}|p{1.9cm}|p{1.9cm}|}
\hline
\multicolumn{4}{|c|}{Exp2: MLP a01b01-unbalanced Acc,Spec,Sens} \\
\hline
&Acc & Spec &Sens  \\
\hline
Baseline&94.22$\pm$0.27&99.44$\pm$0.09&89.12$\pm$0.44 \\
Exp2:Augm&96.70$\pm$0.31 &98.82$\pm$0.24&94.62$\pm$0.51  \\
\hline
\end{tabular}
}
\label{table:ResUnbalanced2}
\vskip -0.4cm
\end{table}
\par \textbf{Results:} The results are shown in Tables 
\ref{table:ResUnbalanced} and \ref{table:ResUnbalanced2}. For Exp2:Augm 
we train the front-end classifiers with AD (i.e., apneic SD and 
RD$_{TRAIN}$ without a04), and for the Baseline we train with 
RD$_{TRAIN}$ without a04. In both cases we evaluate on RD$_{TEST}$.

\par Compared to the Baseline, a clear performance improvement occurs 
for Exp2: Augm. This can be noticed both in terms of accuracy for the MLP 
(Table \ref{table:ResUnbalanced2}, first column) and in terms of kappa 
for all front-end classifiers (all columns of Table 
\ref{table:ResUnbalanced}) . The SVM seems to benefit the most from the 
rebalancing process. Again, in terms of specificity and sensitivity we 
notice a similar behaviour as in the previous experiment with an increase 
in sensitivity and relatively stable specificity.

\par \textbf{Implications for OSA Detection:} As mentioned, OSA data are generally very unbalanced towards non-apneic events. This experiment implies that GAN augmentation with synthetic data can be used to  efficiently rebalance OSA data. This has a positive effect on the  detection of apneic events and  on the overall classification performance for OSA detection, based on the classifiers we experimented with.

\subsection{Experiment3: Personalization with Multiple GANs}

\par In this experiment, the goal is to investigate whether we can 
improve performance by indirect personalization during GAN training.
By \textit{personalization} we mean that we aim to make the learned 
distribution of the GAN we use to generate SD to approach the specific 
distribution of the RD$_{TEST}$ for a given proximity metric (MMD). 
Since we do not use a01b01 for the training of the GAN the method we 
apply is indirect. We 
use two recordings from Apnea-ECG  as RD$_{TEST}$ 
(i.e., a01b01). 
\par
\begin{figure}
  \vskip -0.3cm
\centering

  \includegraphics[width=6cm,height=1.5cm]{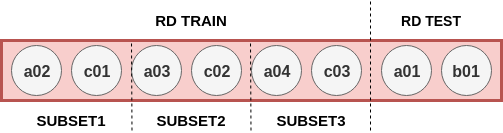}
\caption{Training and Test  sets for Experiment 3}
  \label{fig:Fig_Pers1}
  \vskip -0.3cm
\end{figure}

 \textbf{Experiment Description:}  Based on the discussion of Section 
\ref{meth_exp3}, we separate our training recordings into three subsets 
 (Figure \ref{fig:Fig_Pers1}). Then we create three GANs 
(GAN1, GAN2, and GAN3) and we use each subset to train the respective 
GAN, with a non-zero probability of choosing another subset for the 
gradient update based on a weighted dice toss (see Section 
\ref{meth_exp3}).  We set $p=0.4$ (see Figure \ref{fig:Exp3_1}), i.e., 
for one gradient update of GAN1, the mini-batch is selected with 
probability 0.4 from Subset1, and  probability 0.3 from Subset 2 and 3. 
We do the same  for GAN 2 and 3. The choice of $p$ is made via 
experimental evaluation.

\par Proposition 1 implies that through this training, a GAN converges 
to a mixture of distributions with weights for each subset distribution 
j equal to $P(J=j)$ (see Eq. \ref{gan_value_mixture}). By controlling 
$P(J=j)$ we control the weights of the mixture, and thus the degree to 
which each subset of recordings is represented in SD.
\par We use the validation set from a01b01 (obtained as in Experiment 2) 
for two purposes: (1) to evaluate the SD from the three GANs (SD1, SD2 
and SD3) and (2) to calculate the MMD between SD1-3 and this validation 
set. Then we examine two cases: In Exp3:Augm, SD1, SD2, and SD3 are 
combined with RD$_{TRAIN}$ to form AD. SD1, SD2, and SD3  combined have 
the same size as RD$_{TRAIN}$. In Exp3:AugmP, we identify the SD that 
has the lowest MMD in relation to the validation set, and use the 
corresponding GANi to generate more data until SDi has the size of 
RD$_{TRAIN}$. AD is formed by combining   RD$_{TRAIN}$ and SDi.  In Exp3:AugmP we perform indirect 
personalization, since the SDi selected originates from the GAN that 
best matches the distribution of the subset a01b01, i.e., RD$_{TEST}$ 
based on the MMD metric. This occurs since the validation set is also 
extracted from a01b01. This experiment is also repeated 5 times.

\begin{table}
\vskip -0.5cm
\caption{ Kappa statistic  for front-end classifiers }
\centering
\resizebox{10cm}{!}{%
\begin{tabular}{| p{2.1cm} |p{2cm}|p{2cm}|p{2cm}|p{2cm}|}
\hline
\multicolumn{5}{|c|}{Exp3: Kappa statistic (X$\cdot 10^{-2}$), a01b01 as RD$_{TEST}$ } \\
\hline
&MLP & RF &KNN & SVM \\
\hline
Baseline&92.36$\pm$0.37 &92.88$\pm$0.38&93.12$\pm$0.21&88.20$\pm$0.37\\
Exp3:Augm&93.08$\pm$0.59 &93.6$\pm$0.62  &94.50$\pm$0.39
& 91.72$\pm$0.94\\
Exp3:AugmP&93.36$\pm$0.40&94.36$\pm$0.31 &94.58$\pm$0.17&93.92$\pm$0.23 
\\
\hline
\end{tabular}
}
\label{table:ResPersonalized}
\vskip -0.5cm
\end{table}

\textbf{Results: } The 
results are found in Tables \ref{table:ResPersonalized} and 
\ref{table:ResPersonalized2}. We see that the general behavior is similar to the 
previous experiments. Again there are improvements for the augmented 
cases in relation to the Baseline. There are improvements in sensitivity 
and a small drop in specificity for the MLP cases, which is the case 
also for the other classifiers (with the exception of RF).
\par Generally, Exp3:AugmP, exhibits slightly better performance both in 
terms of kappa  and accuracy. SVM and RF seem to gain the most benefits 
from this approach. Interestingly, in Exp3:AugmP  SVM surpasses MLP in 
terms of kappa.

\begin{table}
\vskip -0.4cm
\caption{ Accuracy, specificity and sensitivity for 
MLP }
\centering
\resizebox{8cm}{!}{%
\begin{tabular}{| p{2cm} |p{1.9cm}|p{1.9cm}|p{1.9cm}|}
\hline
\multicolumn{4}{|c|}{Exp3: MLP a01b01 Acc,Spec,Sens} \\
\hline
&Acc & Spec &Sens  \\
\hline
Baseline&96.18$\pm$0.18&98.92$\pm$0.07& 93.54$\pm$0.25 \\
Exp3:Augm&96.54$\pm$0.29&98.4$\pm$0.19& 94.74$\pm$0.51 \\
Exp3:AugmP&96.68$\pm$0.20&98.64$\pm$0.18& 95.2$\pm$0.25 \\
\hline
\end{tabular}
}
\label{table:ResPersonalized2}
\vskip -0.5cm
\end{table}

\par Also,  to further investigate the viability of Exp3:AugmP method we examine in the Appendix  different recording combinations as RD$_{TEST}$ (i.e.,  a02c01, a04b01 and a03b01) and perform Baseline and Exp3:AugmP  evaluations for the front-end classifiers.  Intriguingly,  for all cases, for all front-end classifiers we notice improvements for the kappa statistic, that vary from (RF, a02c01):0.28$\cdot 10^{-2}$  to (MLP, a03b01): 27.12$\cdot 10^{-2}$, especially for low performing cases e.g., for the (MLP, a03b01) case Baseline kappa is 57.4$\cdot 10^{-2}$ and Exp3:AugmP kappa is 84.5$\cdot 10^{-2}$.
\par \textbf{Implications for OSA Detection:}  This experiment implies that personalization can indeed have a positive impact on classification performance for the detection of OSA. Even the simple indirect approach of Exp3:AugmP exhibits performance advantages for all front-end classifiers in relation to when it is not applied in Exp3:Augm.

\section{Conclusion}
In this work we examined how   dataset augmentation via the use of the  GAN framework can improve the classification performance in three  scenarios for  OSA detection.
We notice  that for all the cases the augmentation 
clearly helps the classifiers to generalize better. Even for the simpler 
classifiers like KNN, we see that augmentation has a beneficial effect 
on performance. The largest performance improvement is achieved for the 
SVM for Experiment 2, and in all the cases the metric that increases the 
most is sensitivity. This leads us to believe that the class balancing 
that GAN can provide with synthetic data can be  useful in 
situations for which one class is much less represented than others. 
This is even more pronounced in cases like OSA detection where the vast 
majority of the data belongs to one of two classes.
\par As a next step we plan to investigate the viability of creating 
synthetic datasets that are differentially private. As health data are 
in many cases withheld from public access, we want to
investigate the performance of front-end classifiers when using 
synthetic datasets that have privacy guarantees and examine how 
this impacts the performance of the classifiers.

\medskip

\bibliographystyle{splncs04}

\bibliography{mybibliography}

\begin{thebibliography}{10}
\providecommand{\url}[1]{\texttt{#1}}
\providecommand{\urlprefix}{URL }
\providecommand{\doi}[1]{https://doi.org/#1}

\bibitem{ApneaEcg}
\url{https://physionet.org/physiobank/database/apnea-ecg/} (1999), [Online;
  accessed 26-3-2019]

\bibitem{MITBIH}
\url{https://physionet.org/physiobank/database/slpdb/} (1999), [Online;
  accessed 26-3-2019]

\bibitem{borji2019pros}
Borji, A.: Pros and cons of gan evaluation measures. Computer Vision and Image
  Understanding  \textbf{179},  41--65 (2019)

\bibitem{che2017boosting}
Che, Z., Cheng, Y., Zhai, S., Sun, Z., Liu, Y.: Boosting deep learning risk
  prediction with generative adversarial networks for electronic health
  records. In: 2017 IEEE International Conference on Data Mining (ICDM). pp.
  787--792. IEEE (2017)

\bibitem{choi2017generating}
Choi, E., Biswal, S., Malin, B., Duke, J., Stewart, W.F., Sun, J.: Generating
  multi-label discrete patient records using generative adversarial networks.
  arXiv preprint arXiv:1703.06490  (2017)

\bibitem{cohen1960coefficient}
Cohen, J.: A coefficient of agreement for nominal scales. Educational and
  psychological measurement  \textbf{20}(1),  37--46 (1960)

\bibitem{douzas2018effective}
Douzas, G., Bacao, F.: Effective data generation for imbalanced learning using
  conditional generative adversarial networks. Expert Systems with applications
   \textbf{91},  464--471 (2018)

\bibitem{durugkar2016generative}
Durugkar, I., Gemp, I., Mahadevan, S.: Generative multi-adversarial networks.
  arXiv preprint arXiv:1611.01673  (2016)

\bibitem{esteban2017real}
Esteban, C., Hyland, S.L., R{\"a}tsch, G.: Real-valued (medical) time series
  generation with recurrent conditional gans. arXiv preprint arXiv:1706.02633
  (2017)

\bibitem{goodfellow2016deep}
Goodfellow, I., Bengio, Y., Courville, A.: Deep learning. MIT press (2016)

\bibitem{goodfellow2014generative}
Goodfellow, I., Pouget-Abadie, J., Mirza, M., Xu, B., Warde-Farley, D., Ozair,
  S., Courville, A., Bengio, Y.: Generative adversarial nets. In: Advances in
  neural information processing systems. pp. 2672--2680 (2014)

\bibitem{gretton2007kernel}
Gretton, A., Borgwardt, K., Rasch, M., Sch{\"o}lkopf, B., Smola, A.J.: A kernel
  method for the two-sample-problem. In: Advances in neural information
  processing systems. pp. 513--520 (2007)

\bibitem{hoang2017multi}
Hoang, Q., Nguyen, T.D., Le, T., Phung, D.: Multi-generator generative
  adversarial nets. arXiv preprint arXiv:1708.02556  (2017)

\bibitem{hochreiter1997long}
Hochreiter, S., Schmidhuber, J.: Long short-term memory. Neural computation
  \textbf{9}(8),  1735--1780 (1997)

\bibitem{hwang2017disease}
Hwang, U., Choi, S., Yoon, S.: Disease prediction from electronic health
  records using generative adversarial networks. arXiv preprint
  arXiv:1711.04126  (2017)

\bibitem{isola2017image}
Isola, P., Zhu, J.Y., Zhou, T., Efros, A.A.: Image-to-image translation with
  conditional adversarial networks. In: Proceedings of the IEEE conference on
  computer vision and pattern recognition. pp. 1125--1134 (2017)

\bibitem{kristiansen2018data}
Kristiansen, S., Hugaas, M.S., Goebel, V., Plagemann, T., Nikolaidis, K.,
  Liest{\o}l, K.: Data mining for patient friendly apnea detection. IEEE Access
   \textbf{6},  74598--74615 (2018)

\bibitem{loberg2018quantifying}
L{\o}berg, F., Goebel, V., Plagemann, T.: Quantifying the signal quality of
  low-cost respiratory effort sensors for sleep apnea monitoring. In:
  Proceedings of the 3rd International Workshop on Multimedia for Personal
  Health and Health Care. pp. 3--11. ACM (2018)

\bibitem{mariani2018bagan}
Mariani, G., Scheidegger, F., Istrate, R., Bekas, C., Malossi, C.: Bagan: Data
  augmentation with balancing gan. arXiv preprint arXiv:1803.09655  (2018)

\bibitem{mirza2014conditional}
Mirza, M., Osindero, S.: Conditional generative adversarial nets. arXiv
  preprint arXiv:1411.1784  (2014)

\bibitem{mogren2016c}
Mogren, O.: C-rnn-gan: Continuous recurrent neural networks with adversarial
  training. arXiv preprint arXiv:1611.09904  (2016)

\bibitem{radford2015unsupervised}
Radford, A., Metz, L., Chintala, S.: Unsupervised representation learning with
  deep convolutional generative adversarial networks. arXiv preprint
  arXiv:1511.06434  (2015)

\bibitem{ravi2017deep}
Rav{\`\i}, D., Wong, C., Deligianni, F., Berthelot, M., Andreu-Perez, J., Lo,
  B., Yang, G.Z.: Deep learning for health informatics. IEEE journal of
  biomedical and health informatics  \textbf{21}(1),  4--21 (2017)

\bibitem{rezaei2019recurrent}
Rezaei, M., Yang, H., Meinel, C.: Recurrent generative adversarial network for
  learning imbalanced medical image semantic segmentation. Multimedia Tools and
  Applications pp. 1--20 (2019)

\bibitem{salimans2016improved}
Salimans, T., Goodfellow, I., Zaremba, W., Cheung, V., Radford, A., Chen, X.:
  Improved techniques for training gans. In: Advances in neural information
  processing systems. pp. 2234--2242 (2016)

\bibitem{sutherland2016generative}
Sutherland, D.J., Tung, H.Y., Strathmann, H., De, S., Ramdas, A., Smola, A.,
  Gretton, A.: Generative models and model criticism via optimized maximum mean
  discrepancy. arXiv preprint arXiv:1611.04488  (2016)

\bibitem{theis2015note}
Theis, L., Oord, A.v.d., Bethge, M.: A note on the evaluation of generative
  models. arXiv preprint arXiv:1511.01844  (2015)

\bibitem{tolstikhin2017adagan}
Tolstikhin, I.O., Gelly, S., Bousquet, O., Simon-Gabriel, C.J., Sch{\"o}lkopf,
  B.: Adagan: Boosting generative models. In: Advances in Neural Information
  Processing Systems. pp. 5424--5433 (2017)

\bibitem{traaen2019treatment}
Traaen, G.M., Aaker{\o}y, L., et~al.: Treatment of sleep apnea in patients with
  paroxysmal atrial fibrillation: Design and rationale of a randomized
  controlled trial. Scandinavian Cardiovascular Journal (52:6,pp. 372-377),
  1--20 (January 2019)

\bibitem{yu2017seqgan}
Yu, L., Zhang, W., Wang, J., Yu, Y.: Seqgan: Sequence generative adversarial
  nets with policy gradient. In: Thirty-First AAAI Conference on Artificial
  Intelligence (2017)

\end{thebibliography}

\appendix

\section{Classifier Parameters}
Appendix A summarizes the parameters and details used for the front-end clas-
sifiers and GAN.
\subsection{Front-End Classifier Parameters}

 As mentioned we use SVM, KNN, MLP and RF as our front-end classifiers. The parameters we use are:  

\begin{itemize}
\item MLP: We use a small feed-forward neural network with one hidden layer with 100 neurons, adam optimizer, relu activation, learning rate equal to 0.001, a batch size of 200, no regularization,   and the other parameters set on the default values based on the implementation from https://scikit-learn.org/stable/.
\item KNN:  K-Nearest Neighbor with five neighbors, euclidean distance, weights based on the distance from the target and the other parameters set on the default values based on the implementation from https://scikit-learn.org/stable/.
\item SVM: A Support Vector Machine with an Radial Basis Function kernel, penalty parameter of the error term equal to 1,  and the other parameters set on the default values s based on the implementation from https://scikit-learn.org/stable/.
\item RF: Random forest comprised of 50 trees, Gini impurity as function to measure the quality of the split, and the other parameters set on the default values s based on the implementation from https://scikit-learn.org/stable/.
\end{itemize}

\subsection{GAN LSTM}

We use TensorFlow and implement two conditional LSTMs one as generator that takes input from a normal distribution with mean=0 and std=1 and outputs sequences of NAF data and a discriminator LSTM that takes as input real and synthetic NAF inputs and outputs D(x) that estimates the chance that the input is real. The inputs, and thus the updates for both nets are per-sample and not per minute. As input for G and D we use an additional conditional vector that maps non-apneas as zero and apneas as one (again per sample). G gradient updates are performed via standard gradient descent with 0.01 learning rate and D via adam optimizer with learning rate 0.01. The mini-batch size is 50, the size (hidden units) of D and G is 300. All these values correspond to the most usual cases, but different configurations have been tested.

\section{ Accuracy, Sensitivity, and Specificity of RF, KNN, and SVM for Exp1}

Appendix B complements the results from Experiment 1 presented in the paper with the accuracy, specificity, and sensitivity of RF, KNN; and SVM.

\begin{table}
\centering
\resizebox{7cm}{!}{%
\begin{tabular}{ | p{2cm} |p{1.9cm}|p{1.9cm}|p{1.9cm}|}
\hline
\multicolumn{4}{|c|}{RF Classifier Apnea-ECG} \\
\hline
&Acc & Spec &Sens  \\
\hline
Baseline&95.18$\pm$0.12&95.77$\pm$0.23&94.32$\pm$0.40\\
Exp1:Synth&92.19$\pm$0.27&92.53$\pm$0.75&92.46$\pm$0.83\\
Exp1:Augm &95.52$\pm$0.13&95.38$\pm$0.22&95.74$\pm$0.22\\
\hline
\end{tabular}
}

\centering
\resizebox{7cm}{!}{%
\begin{tabular}{ | p{2cm} |p{1.9cm}|p{1.9cm}|p{1.9cm}|}
\hline
\multicolumn{4}{|c|}{RF Classifier MIT-BIH} \\
\hline
&Acc & Spec &Sens  \\
\hline
Baseline&68.57$\pm$0.54&87.43$\pm$0.51.9&41.57$\pm$0.98\\
Exp1:Synth&61.97$\pm$0.51&66.15$\pm$1.65&55.8$\pm$2.21 \\
Exp1:Augm &68.02$\pm$0.39&76.22$\pm$1.61&56.91$\pm$2.36\\
\hline
\end{tabular}
}
 \caption{Accuracy, specificity, and sensitivity for the RF classifier }

\end{table}

\begin{table}

\centering
\resizebox{7cm}{!}{%
\begin{tabular}{ | p{2cm} |p{1.9cm}|p{1.9cm}|p{1.9cm}|}
\hline
\multicolumn{4}{|c|}{KNN Classifier Apnea-ECG} \\
\hline
&Acc & Spec &Sens  \\
\hline
Baseline&94.34$\pm$0.20&96.88$\pm$0.20&90.94$\pm$0.39\\
Exp1:Synth&93.07$\pm$0.24&92.27$\pm$0.62&94.2$\pm$0.56 \\
Exp1:Augm &95.20$\pm$0.17&96.01$\pm$0.31&94.04$\pm$0.51\\
\hline
\end{tabular}
}
\resizebox{7cm}{!}{%
\begin{tabular}{ | p{2cm} |p{1.9cm}|p{1.9cm}|p{1.9cm}|}
\hline
\multicolumn{4}{|c|}{KNN Classifier MIT-BIH} \\
\hline
&Acc & Spec &Sens  \\
\hline
Baseline&65.37$\pm$0.52&74.43$\pm$0.49&53.31$\pm$1.22\\
Exp1:Synth&57.20$\pm$0.69&50.78$\pm$3.06&66.99$\pm$3.82\\
Exp1:Augm &64.99$\pm$0.51&65.26$\pm$0.83&64.64$\pm$1.24\\
\hline
\end{tabular}
}
 \caption{Accuracy, specificity, and sensitivity for the KNN classifier }

 \end{table}

\begin{table}

\centering
\resizebox{7cm}{!}{%
\begin{tabular}{ | p{2cm} |p{1.9cm}|p{1.9cm}|p{1.9cm}|}
\hline
\multicolumn{4}{|c|}{SVM Classifier Apnea-ECG} \\
\hline
&Acc & Spec &Sens  \\
\hline
Baseline&87.38$\pm$0.21&81.94$\pm$0.43&95.35$\pm$0.25\\
Exp1:Synth&87.48$\pm$0.27&80.78$\pm$0.40&97.5$\pm$0.45 \\
Exp1:Augm &88.40$\pm$0.29&82.13$\pm$0.61&97.53$\pm$0.61\\
\hline
\end{tabular}
}
\resizebox{7cm}{!}{%
\begin{tabular}{ | p{2cm} |p{1.9cm}|p{1.9cm}|p{1.9cm}|}
\hline
\multicolumn{4}{|c|}{SVM Classifier MIT-BIH} \\
\hline
&Acc & Spec &Sens  \\
\hline
Baseline&59.11$\pm$0.56&100$\pm$0.0&0.0$\pm$0.00\\
Exp1:Synth&57.2$\pm$0.69&50.78$\pm$3.06&66.99$\pm$3.82\\
Exp1:Augm &57.75$\pm$0.63&57.80$\pm$1.68&57.62$\pm$2.07\\
\hline
\end{tabular}
}
 \caption{Accuracy, specificity, and sensitivity for the SVM classifier }
 \end{table}

\section{ Accuracy, Sensitivity, and Specificity of RF, KNN, and SVM for Exp2}

Appendix C complements the results from Experiment 2 presented in the paper with accuracy, sensitivity, and specificity of RF, KNN, and SVM.

\begin{table}[H]
 \vskip -0.3cm
\centering
\resizebox{7cm}{!}{%
\begin{tabular}{ | p{2cm} |p{1.9cm}|p{1.9cm}|p{1.9cm}|}
\hline
\multicolumn{4}{|c|}{RF Classifier Apnea-ECG} \\
\hline
&Acc & Spec &Sens  \\
\hline
Baseline&95.96$\pm$0.13&99.22$\pm$0.11&92.78$\pm$0.28\\
Exp2:Augm &97.28$\pm$0.08&98.96$\pm$0.20&95.62$\pm$0.19\\
\hline
\end{tabular}
}
\centering
\resizebox{7cm}{!}{%
\begin{tabular}{ | p{2cm} |p{1.9cm}|p{1.9cm}|p{1.9cm}|}
\hline
\multicolumn{4}{|c|}{SVM Classifier Apnea-ECG} \\
\hline
&Acc & Spec &Sens  \\
\hline
Baseline&87.3$\pm$0.11&99.12$\pm$0.08&75.74$\pm$0.37\\
Exp2:Augm &96.44$\pm$0.32&98.6$\pm$0.10&94.28$\pm$0.75\\
\hline
\end{tabular}
}

\centering
\resizebox{7cm}{!}{%
\begin{tabular}{ | p{2cm} |p{1.9cm}|p{1.9cm}|p{1.9cm}|}
\hline
\multicolumn{4}{|c|}{KNN Classifier Apnea-ECG} \\
\hline
&Acc & Spec &Sens  \\
\hline
Baseline&96.58$\pm$0.08&99.28$\pm$0.08&92.8$\pm$1.15\\
Exp2:Augm &97.38$\pm$0.22&99.06$\pm$0.08&95.82$\pm$0.31\\
\hline
\end{tabular}
}
 \caption{Accuracy, specificity, and sensitivity for the RF, KNN, and SVM classifiers }

\end{table}

\section{ Accuracy Sensitivity Specificity of RF,KNN,SVM for Exp3}

Appendix D complements the results from Experiment 3 presented in the paper with accuracy, sensitivity, and specificity of RF, KNN, and SVM.

\begin{table}[H]
 \vskip -0.3cm
\centering
\resizebox{7cm}{!}{%
\begin{tabular}{ | p{2cm} |p{1.9cm}|p{1.9cm}|p{1.9cm}|}
\hline
\multicolumn{4}{|c|}{RF Classifier Apnea-ECG} \\
\hline
&Acc & Spec &Sens  \\
\hline
Baseline&96.44$\pm$0.19&96.94$\pm$0.33&95.98$\pm$0.25\\
Exp3:Augm &96.8$\pm$0.31&98.06$\pm$0.51&95.54$\pm$0.17\\
Exp3:AugmP &97.16$\pm$0.15&98.98$\pm$0.15&95.40$\pm$0.17\\
\hline
\end{tabular}
}

\centering
\resizebox{7cm}{!}{%
\begin{tabular}{ | p{2cm} |p{1.9cm}|p{1.9cm}|p{1.9cm}|}
\hline
\multicolumn{4}{|c|}{SVM Classifier Apnea-ECG} \\
\hline
&Acc & Spec &Sens  \\
\hline
Baseline&94.12$\pm$0.18&98.76$\pm$0.06&89.64$\pm$0.36\\
Exp3:Augm &95.86$\pm$0.47&97.48$\pm$0.73&94.34$\pm$0.46\\
Exp3:AugmP &96.96$\pm$0.11&98.40$\pm$0.05&95.62$\pm$0.21\\
\hline
\end{tabular}
}

\centering
\resizebox{7cm}{!}{%
\begin{tabular}{ | p{2cm} |p{1.9cm}|p{1.9cm}|p{1.9cm}|}
\hline
\multicolumn{4}{|c|}{KNN Classifier Apnea-ECG} \\
\hline
&Acc & Spec &Sens  \\
\hline
Baseline&96.56$\pm$0.08&99.06$\pm$0.08&94.10$\pm$0.13\\
Exp3:Augm &97.29$\pm$0.15&99.25$\pm$0.16&95.43$\pm$0.27\\
Exp3:AugmP &97.29$\pm$0.08&99.11$\pm$0.08&95.52$\pm$0.13\\
\hline
\end{tabular}
}
 \caption{Accuracy, specificity, and sensitivity for the RF, KNN and SVM classifiers }

\end{table}

\section{Experiment 3: Results of additional recording combinations for $RD_{TEST}$}

Appendix E supplements the results of Experiment 3 in the paper for kappa with the additional recording combinations a02c01, a03b01,a04b01 for RD$_{TEST}$.

\begin{table}[H]
 \vskip -0.3cm

\centering
\resizebox{7.5cm}{!}{%
\begin{tabular}{ | p{2cm} |p{1.9cm}|p{1.9cm}|p{1.9cm}|p{1.9cm}|}
\hline
\multicolumn{5}{|c|}{Kappa statistic (X$\cdot10^{-2}$) combination: a02c01} \\
\hline
&MLP & RF & KNN & SVM  \\
\hline
Baseline&80.68$\pm$1.0&91.68$\pm$0.39&80.83$\pm$0.49&87.32$\pm$0.50\\
Exp3:AugmP &86.26$\pm$1.39&91.96$\pm$0.28&81.72$\pm$0.70&92.97$\pm$0.35\\
\hline
\end{tabular}
}

\centering
\resizebox{7.5cm}{!}{%
\begin{tabular}{ | p{2cm} |p{1.9cm}|p{1.9cm}|p{1.9cm}|p{1.9cm}|}
\hline
\multicolumn{5}{|c|}{Kappa statistic (X$\cdot10^{-2}$) combination: a04b01} \\
\hline
&MLP & RF & KNN & SVM  \\
\hline
Baseline&54.45$\pm$0.68&56.46$\pm$1.37&71.77$\pm$1.08&81.35$\pm$0.37\\
Exp3:AugmP &71.17$\pm$3.04&83.19$\pm$1.39&83.35$\pm$0.49&92.51$\pm$0.14\\
\hline
\end{tabular}
}

\centering
\resizebox{7.5cm}{!}{%
\begin{tabular}{ | p{2cm} |p{1.9cm}|p{1.9cm}|p{1.9cm}|p{1.9cm}|}
\hline
\multicolumn{5}{|c|}{Kappa statistic  (X$\cdot10^{-2}$) combination: a03b01} \\
\hline
&MLP & RF & KNN & SVM  \\
\hline
Baseline&57.41$\pm$1.0&59.78$\pm$0.79&41.41$\pm$0.39&83.57$\pm$0.49\\
Exp3:AugmP &84.57$\pm$0.26&78.02$\pm$1.9&50.41$\pm$4.61&87.34$\pm$0.73\\
\hline
\end{tabular}
}
 \caption{Kappa for different RD$_{TEST}$ combinations a02c01, a03b01, a04b01 }

\end{table}

It is worth to mention that these are all the combinations we examined. No additional combinations were examined.

\section{Reasons for the design of the T-Metric}
Appendix F gives a detailed explanation for our choice of the T-metric.

The fundamental observation is that the classifiers have to determine the underlying
training set distribution. If the trained classifiers perform well on the test set, then the train
and test distributions should be similar.

We follow the evaluation approaches presented in [1] called 'Train on Synthetic-Test on Real'
(TSTR), and 'Train on Real-Test on Synthetic' (TRTS). The classifiers are trained on
apnea classification with a synthetic dataset and tested on the real training data to perform the
TSTR test, and the opposite procedure is performed for the TRTS test.
As mentioned in [1], TSTR individually is a potentially better metric for similarity than
TRTS, as it is sensitive to mode collapse. If a classifier is trained on synthetic
data which have many collapsed modes, the performance on the real training data would be
low. However, we argue that individually both metrics can be problematic for distribution
comparison in certain cases, such as in binary classification. For example, if the
synthetic distribution has a larger difference in variance between the classes, TSTR will not capture this,
whereas TRTS will and vice versa. By including both measures in the metric 
 this issue gets mostly solved, since the metric becomes sensitive to this
variance. Figure \ref{fig1} illustrates a concrete example in which the data points from the synthetic distribution
are depicted with magenta and cyanic, and from the real with red and blue.
In the TSTR test, the classifier learns the magenta separation hyperplane, and in the TRTS the blue
separation hyperplane. Here TRTS captures better the dissimilarity between real and
synthetic data. The opposite holds if the real and the synthetic distributions are swapped in the example.
 
\begin{figure*}
\centering
\includegraphics[scale=0.45]{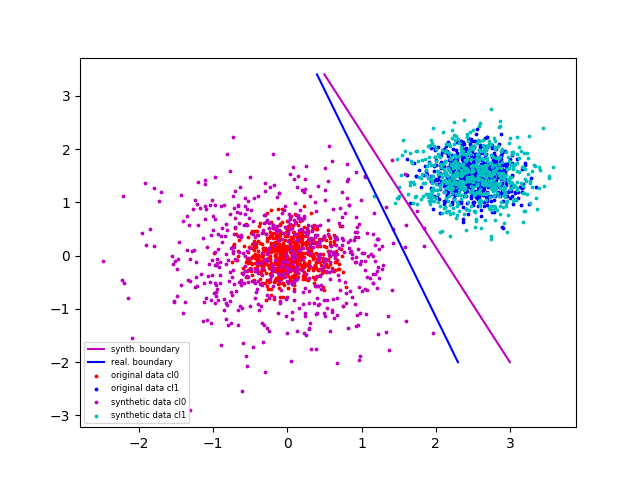}
\caption{Example of TSTR and TRTS issues} \label{fig1}
\end{figure*}
\par The TRTS and TSTR metrics are combined via using the
harmonic mean of the two measurements:
\begin{equation}
T=\frac{2*TSTR*TRTS}{TSTR+TRTS}
\end{equation}
 The harmonic mean was chosen instead of other potential measures (e.g., average) for
several reasons. First, the harmonic mean is punishing more the differences of the scores, so for example if TSTR is 0.5 and TRTS is 1, the average would be 0.75, whereas the harmonic
mean would be 0.66. In order to measure similarity, both scores should be high, so this is a
desired property. Additionally, if mode collapse occurs, TRTS is expected to have a high performance whereas TSTR is significantly lower. An average would not be able to capture that problem
well. Finally, it is worth noting
that the T-metric is sensitive to the classifiers’ capacity, and this is why all four
classifiers are used for the test. Additionally, which metric is used for the TSTR and TRTS tests plays also an important role. We experiment with accuracy and AUROC

\par A potential problem of this method is that the separation hyperplane criterion could
be insensitive to the spread towards unimportant directions for the classification in the
feature space. 

\section{Proposition 1}

Appendix G gives a detailed description of Proposition 1.
Let X random variable (r.v) such that:

\begin{equation}
X=\begin{cases} 
      X_0 \sim P_{rec_0} &,if A_0 \\
      X_1 \sim P_{rec_1} &,if A_1 \\
     ...\\
      X_N \sim P_{rec_N} &,if A_N \\
   \end{cases}
\end{equation}

where:
\\\\
$A_i$:\textit{The event that  r.v $Z$=i with $Z \sim$ categorical distribution with $P(A_i)=w_i$}  
\\\\
 Then:

\begin{theorem} 
$X \sim P_s$ with PDF $p_s(\mathbf{x})= \sum_i^{N}w_ip_{rec_i}(\mathbf{x})$  where $p_{rec_i}$ the PDF of  $P_{rec_i}, i=...N$ 
\end{theorem} 

We have $\forall$ subset $S$ of the feature space $D$ ($S \subseteq D$), from the Bayes rule:

\begin{align*}
P(X \in S)=&\sum_iP(X\in S|A_i)P(A_i) \\
=& \sum_i P(X_i \in S)P(A_i)\\
=& \sum_i w_i\int_S p_{rec_i}(\mathbf{x})d\mathbf{x} =\int_S \sum_i w_i p_{rec_i}(\mathbf{x})d\mathbf{x}=\int_S p_s(\mathbf{x})d\mathbf{x}
\end{align*}

$\forall S\subseteq D$, from eq.(1), and since $A_i$ disjoint. So $X\sim P_s$.
\\

We create a dataset $D_x=\{X^{(1)},X^{(2)},...,X^{(m)} \}$ All of the elements of $D_x$, are random variables which  follow Eq. (1), so $X^{(1)},X^{(2)},...,X^{(m)}\sim p_s$. From [11], we follow Algorithm (1) with $D_x$ as the real dataset. Under the given conditions, the generator distributions $p_g$ will converge to the real data distribution $p_{data}$. In our case $p_{data}=p_s$, so from Proposition 2 from [11], $p_g$ converges to $p_s$.
Note that we can control the probabilities $w_i$ which gives us the ability to control the priority of specific recordings in the synthetic dataset. 

\section{Additional Images of Noisy Real and Synthetic Apneic NAF Data}

Appendix H includes two additional images of noisy real and synthetic apneic NAF data (Figure 2). 
\begin{figure*}
\centering
\includegraphics[scale=0.225]{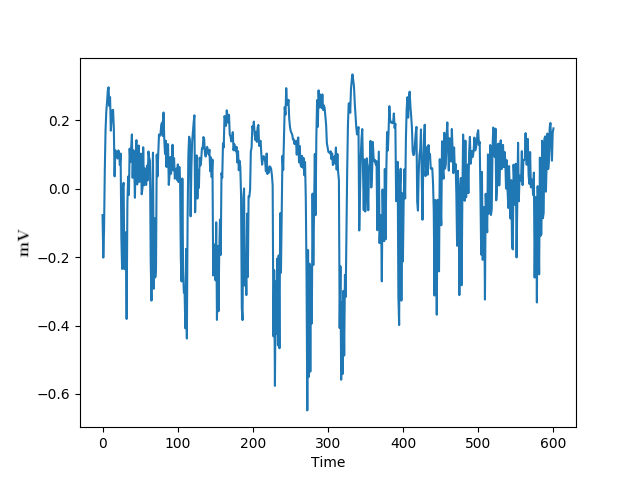}
\includegraphics[scale=0.225]{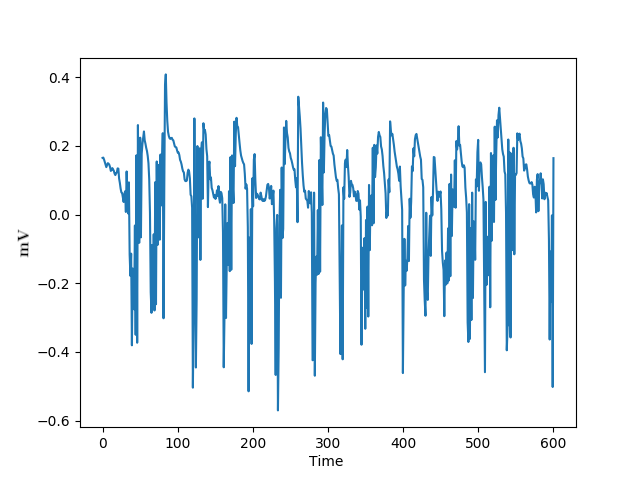}
\label{fig_sk}
\caption{Noisy real apneic data (left) and good noisy synthetic apneic data (right) for 600sec } 
\end{figure*}

\end{document}